%% file: main.tex
\documentclass[]{spie}  

\usepackage{amsmath,amsfonts,amssymb}
\usepackage{graphicx}
\usepackage[colorlinks=true, allcolors=blue]{hyperref}

\usepackage{subfigure}
\usepackage{color}

\title{Revisit Dictionary Learning for Video Compressive Sensing under the Plug-and-Play Framework}

\author[a]{$^{*}$Qing Yang}
\author[b]{$^{*}$Yaping Zhao}
\affil[a]{State Key Laboratory of Precision Spectroscopy, East China Normal University}
\affil[b]{The University of Hong Kong}

\authorinfo{$^{*}$Equal contribution. 
}

\begin{document} 
\maketitle

\begin{abstract}
 Aiming at high-dimensional (HD) data acquisition and analysis, snapshot compressive imaging (SCI) obtains the 2D compressed measurement of HD data with optical imaging systems and reconstructs HD data using compressive sensing algorithms. While the Plug-and-Play (PnP) framework offers an emerging solution to SCI reconstruction, its intrinsic denoising process is still a challenging problem. Unfortunately, existing denoisers in the PnP framework either suffer limited performance or require extensive training data. In this paper, we propose an efficient and effective shallow-learning-based algorithm for video SCI reconstruction. Revisiting dictionary learning methods, we empower the PnP framework with a new 
denoiser, the kernel singular value decomposition (KSVD). Benefited from the advent of KSVD, our algorithm retains a good trade-off among quality, speed, and training difficulty.
 On a variety of datasets, both quantitative and qualitative evaluations of our simulation results demonstrate the effectiveness of our proposed method. In comparison to a typical baseline using total variation, our method achieves around $2$ dB improvement in PSNR and 0.02 in SSIM. We expect that our proposed PnP-KSVD algorithm can serve as a  new baseline for  video SCI reconstruction.
\end{abstract}

\input{intro.tex}
\input{related.tex}
\input{method.tex}

\input{experiment.tex}

\input{conclusion.tex}  

\bibliography{report,reference_sideinfor,reference_ECCV, reference_xin} 
\bibliographystyle{spiebib} 

\end{document}

%% file: intro.tex
\section{Introduction}

The rapid development in machine learning (ML) and artificial intelligence (AI) leads to the necessity and urgency of high-dimensional (HD) visual signal acquisition and analysis. Though optical sensors are diversely growing to capture visual (and other) signals, most of them are composed of two-dimensional (2D) arrays. The fundamental issue remains: how to acquire HD ($\ge3$D) data efficiently. To address this problem, snapshot compressive imaging (SCI) is proposed to obtain the 2D {\em compressed} measurement of HD data with optical imaging systems (encoder), and then reconstruct HD data using compressive sensing algorithms (decoder)~\cite{Yuan2021_SPM}. During the past twenty years, both the hardware~\cite{Qiao2020_APLP,Qiao2020_CACTI} and algorithms~\cite{Ma19ICCV,Cheng20ECCV_BIRNAT} have been improved significantly for SCI.
In this paper, we focus on the algorithmic part, \textit{e.i.}, the decoding part in SCI.

The latest trend in SCI is to exploit the plug-and-play (PnP) framework for HD data reconstruction, which generally leads to a closed-form solution updating sub-problem (projecting the measurement to the signal space) and an image denoising sub-task~\cite{yuan2020plug}. Such a denoising process is an inevitable step under the PnP framework, and therefore has a major effect on the final reconstruction. While the PnP framework has been successfully applied in video SCI, choosing a proper denoiser is still a challenging problem.

\begin{figure}[ht]
    \centering
  \includegraphics[width=\textwidth]{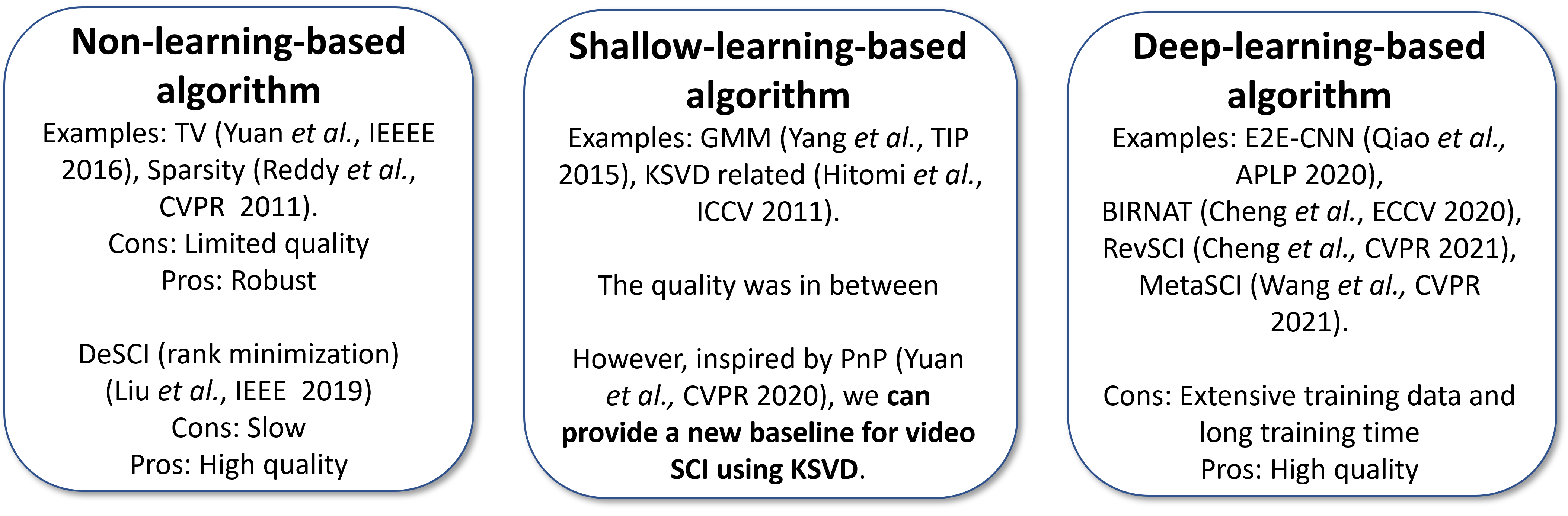}
  \caption{Investigation of non-learning-based algorithm, shallow-learning-based algorithm and deep-learning-based algorithm for SCI reconstruction. Inspired by the PnP framework, a new baseline for video SCI using KSVD is proposed in this paper.}
  \label{fig:gap}
\end{figure}

To perform denoising, various existing algorithms could be categorized into non-learning-based ones, shallow-learning-based ones, and deep-learning-based ones, shown in Figure \ref{fig:gap} . Among non-learning-based algorithms, total variation (TV) \cite{yuan2016generalized} and sparsity-based~\cite{reddy2011p2c2} ones exhibit limited performance, and DeSCI~\cite{Liu18TPAMI} suffers low speed. On the other hand, deep-learning-based ones~\cite{Ma19ICCV,Meng_GAPnet_arxiv2020} need extensive training data and long training time for the sake of high quality~\cite{Cheng20ECCV_BIRNAT,Cheng2021_CVPR_ReverSCI}.

In contrast, shallow-learning-based algorithms stand out because of their good trade-off among quality, speed, and training difficulty. 
This line of research mostly includes the patch-based learning methods such as Gaussian mixture models (GMM)~\cite{Yang14GMM,Yang14GMMonline} and dictionary learning~\cite{Hitomi11ICCV}.
Compared with the state-of-the-art optimization-based algorithm, DeSCI (Decompress-Snapshot-Compressive-Imaging)~\cite{Liu18TPAMI}, GMM can only provide limited performance. Though dictionary learning methods have been used for video SCI in~\cite{Hitomi11ICCV}, the algorithm developed therein is fairly complicated and thus not efficient.   
Thanks to the PnP framework~\cite{Venkatakrishnan_13PnP,Sreehari16PnP,Chan2017PlugandPlayAF}, now various denoisers can be used in an in-situ manner. As mentioned in~\cite{yuan2020plug,Yuan2021_TPAMI_PnP}, DeSCI can be recognized as PnP-WNNM, where WNNM (weighted nuclear norm minimization) is one of the state-of-the-art denoising algorithms~\cite{Gu14CVPR}. However, as mentioned before, DeSCI is very slow, though it can provide high-quality images, and this precludes its wide applications in real SCI systems. 
We aim to address the following question in this paper:
\begin{itemize}
    \item Can we build an efficient and effective shallow-learning-based algorithm for SCI reconstruction?
\end{itemize}

To answer this question, we revisit the renowned dictionary learning method, the kernel singular value decomposition (KSVD)~\cite{Aharon06TSP} and apply it for SCI reconstruction. 
KSVD is much easier to train than convolutional neural networks (CNN), also with the smaller requirement of training data. In addition, for task-specific reconstruction, it is easy to train a different dictionary for different types of data, such as normal images, CT images, microscopy images, diffraction images, \textit{etc}. It has been noticed in~\cite{Zheng20_PRJ_PnP-CASSI} that the CNN based denoiser is not flexible for different tasks. For example, the well-trained FFDnet~\cite{Zhang18TIP_FFDNet} for visual images cannot perform well for hyperspectral image reconstruction. Therefore, a new denoiser has to be re-trained and this takes weeks as well as a large amount of training data and computational resources. 
It is very challenging to train a deep model for various datasets.

Bearing these in mind, in this paper, we apply KSVD based denoiser into the PnP framework and we show that with a well-trained dictionary, the results of our proposed PnP-KSVD are superior to those of GAP-TV.
Specific contributions of our work are listed as follows:
\begin{itemize}
    \item We provide a new baseline for video SCI reconstruction by integrating the KSVD method into the PnP framework.

    \item In comparison to others, our results retain a good trade-off among quality, speed, and training difficulty.
    
    \item Extensive experiments on both simulated and real data demonstrate the robustness of our method.
\end{itemize}

%% file: related.tex
\section{related work}

\subsection{Non-learning-based algorithm}
  Traditional work of non-learning-based algorithms used in SCI usually includes total variation (TV)\cite{yuan2016generalized}, sparsity~\cite{reddy2011p2c2,Yuan14CVPR} and DeSCI\cite{Liu18TPAMI} based on rank minimization. From previous experimental data, TV and sparsity-based algorithms are robust but of limited quality. DeSCI is of high quality but slow. 
 
\subsection{Deep-learning-based algorithm}
As we all know, recently the most popular algorithms in SCI are based on deep learning such as E2E-CNN \cite{Qiao2020_APLP}, BIRNAT \cite{Cheng20ECCV_BIRNAT}, RevSCI\cite{Cheng2021_CVPR_ReverSCI} and MetaSCI \cite{Wang2021_CVPR_MetaSCI}. Complicated deep learning networks perform high-quality reconstruction efficiently but need extensive training data and long training time.

\subsection{Shallow-learning-based algorithm}
The performance of shallow-learning-based algorithms such as GMM\cite{} and KSVD related \cite{Hitomi11ICCV} were between the two algorithms mentioned above. Considering this fact, we provide a new baseline for video SCI using the KSVD algorithm inspired by the PnP~\cite{yuan2020plug} framework. KSVD is a kind of traditional iteration algorithm of dictionary learning in the image or video denoising. In this article, we plug the KSVD algorithm into the PnP framework as denoising for video SCI systems creatively. And the experimental results show pretty well. We hope this study would provide a meaningful inspiration for researchers in promoting the properties of the SCI system. 


%% file: method.tex
\section{Method}
\begin{figure}[htbp]
\centering
\subfigure[]{
\begin{minipage}[t]{0.5\textwidth}
\centering
\includegraphics[width=\textwidth]{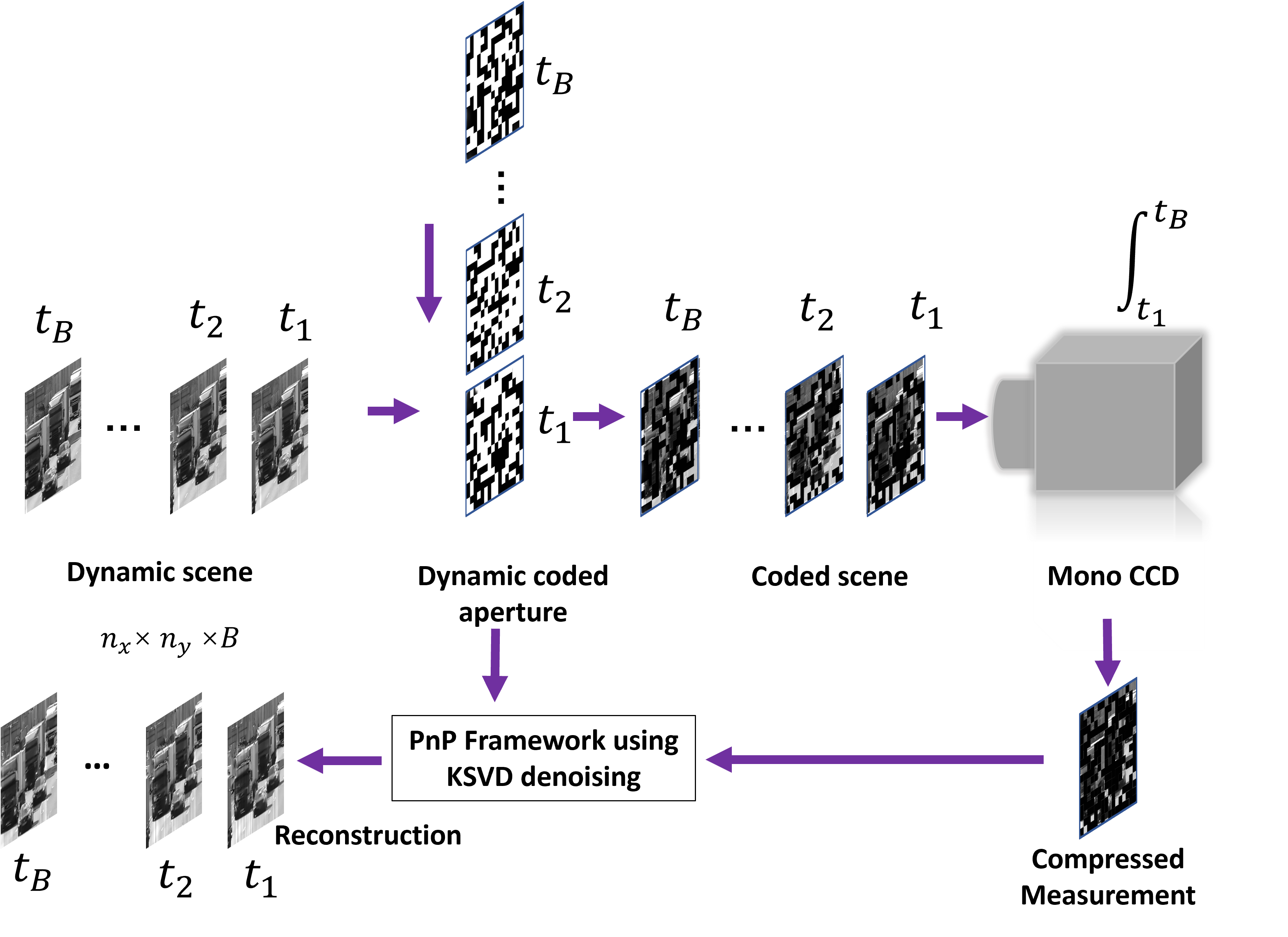}
\label{fig:2a}
\end{minipage}%
}%
\subfigure[]{
\begin{minipage}[t]{0.5\textwidth}
\centering
\includegraphics[width=\textwidth]{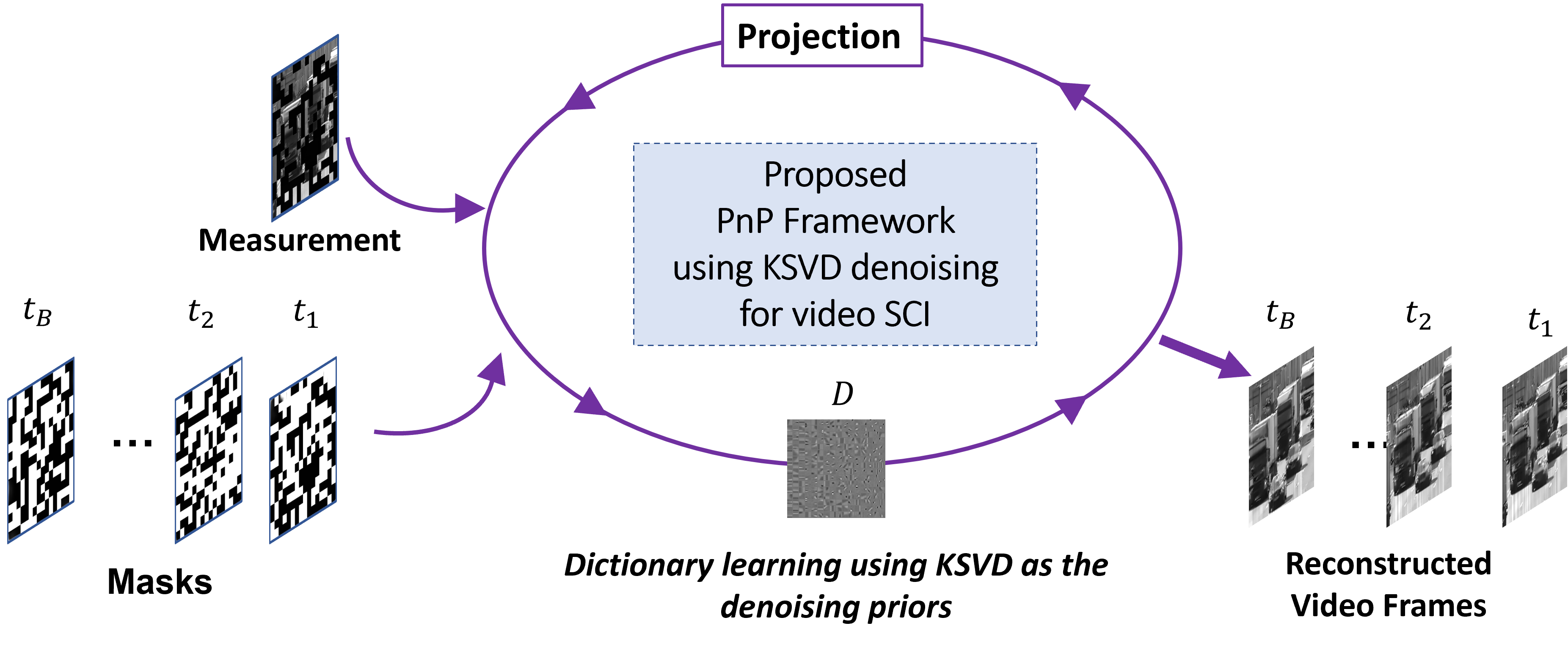}
\label{fig:2b}
\end{minipage}%
}%
\centering
\caption{Principle and  framework of PnP-KSVD for video SCI. (a) Dynamic coded aperture called masks coding dynamic scenes equivalents differential video in temporal dimension. Using 
camera like CCD to integrate coded
scene and then generating compressed measurement (hiding information of 
time dimension). PnP framework helps compressed measurement decode. 
Meanwhile, using algorithm like
KSVD as denoiser to reduce 
inevitable errors. (b) Put 
data of measurement and coded
masks into the PnP-KSVD iterative algorithm. Using dictionary 
$\mathbf{D}$ generated by KSVD 
algorithm as priors to accomplish reconstruction. }
\end{figure}

\subsection{Mathematical Model of SCI}
As Figure \ref{fig:2a} shows, a typical video SCI system compresses $B$ video frames, denoted as $\mathbf{X} \in {\mathbb{R}}^{n_x \times n_y \times B}$, into a 2D measurement $\mathbf{Y} \in {\mathbb{R}}^{n_x \times n_y}$ by coding masks $\mathbf{M} \in {\mathbb{R}}^{n_x \times n_y \times B}$. This process could be formulated as
\begin{align}
    \mathbf{Y} = \sum^B_{b=1} \mathbf{M}_b \odot \mathbf{X}_b +  \mathbf{E},
\label{eq:1}
\end{align}
where ${\mathbb{E}}^{n_x \times n_y }$ denotes the noise; $\mathbf{M}_b = \mathbf{M}(:, :, b)$ and $\mathbf{X}_b = \mathbf{X}(:, :, b) \in {\mathbb{R}}^{n_x \times n_y}$ represent the $b$-th coding mask and the corresponding video frame, respectively; $\odot$ denotes the Hadamard (element-wise) product. Mathematically, the measurement in Equation \eqref{eq:1} can be rewritten as
\begin{align}
    \mathbf{y} = \mathbf{\Phi} \mathbf{x} + \mathbf{e},
    \label{Eq:SCI_forward}
\end{align}
where $\mathbf{y} = \text{Vec}(\mathbf{Y}) \in {\mathbb{R}}^{n_xn_y}$, $\mathbf{e} = \text{Vec}(\mathbf{E}) \in {\mathbb{R}}^{n_x n_y}$,  $\mathbf{x} = \text{Vec}(\mathbf{X}) \in {\mathbb{R}}^{n_x n_y B}$, and thus $\mathbf{\Phi}\in {\mathbb{R}}^{n_x n_y \times n_x n_y B}$.
The forward model of SCI is fairly simple as shown in Eq.~\eqref{Eq:SCI_forward}.
The real challenge is the inverse problem, i.e., the decoder or reconstruction algorithms~\cite{Jalali19TIT_SCI}. 
More specifically, given the compressed measurement $\mathbf{y}$ and sensing matrix $\mathbf{\Phi}$.

\subsection{PnP framework for SCI}
As can be seen from Equation~\eqref{Eq:SCI_forward}, the inverse problem of SCI is ill-posed ($\mathbf{\Phi}$ have more columns than rows) and thus a regularizer is usually employed. 
With a regularization function $R$ that could be applied to $\mathbf{x}$, we could estimate $\mathbf{\hat{x}}$ by solving the optimization problem
\begin{align}
    \mathbf{\hat{x}} = \mathop{\arg\min}_{ \mathbf{x} } \frac{1}{2}|| \mathbf{y} - \mathbf{\Phi} \mathbf{x} ||_2^2 + R(\mathbf{x}),
\label{eq:opt}
\end{align}
where $\mathbf{\hat{x}} \in {\mathbb{R}}^{n_x n_y B}$ is an estimate of the original visual signal $\mathbf{x}$.
Using the alternating direction method of multiplier (ADMM)~\cite{Boyd11ADMM}, the unconstrained optimization in Equation \eqref{eq:opt} can be converted into
\begin{align}
\begin{aligned}
    \label{eq:opt2}
    (\mathbf{\hat{x}}, \mathbf{\hat{v}}) = \mathop{\arg\min}_{\mathbf{x}, \mathbf{v}} \frac{1}{2}|| \mathbf{y} - \mathbf{\Phi} \mathbf{x} ||_2^2 + R(\mathbf{v}), \\
    ~{\text { subject to }}~\ \mathbf{x}=\mathbf{v}.
\end{aligned}
\end{align}
For lower computational workload, we could use the generalized alternating projection (GAP)~\cite{Liao14GAP} as a special case of ADMM. To solve Equation \eqref{eq:opt2}, GAP updates $\mathbf{x}^k$ and $\mathbf{v}^k$ as follows,
\begin{align}
\label{eq:gap-x}
    \mathbf{x}^{(k+1)} &= \mathbf{v}^k + {\mathbf{\Phi}}^\top (\mathbf{\Phi} {\mathbf{\Phi}}^\top)^{-1}(\mathbf{y}-\mathbf{\Phi} \mathbf{v}^k),\\
    \mathbf{v}^{(k+1)} &= \mathcal{D} (\mathbf{x}^{(k+1)}), \label{Eq:denoisePnP}
\end{align}
where $\mathcal{D}$ is the denoiser (or the denoising prior) being used in the PnP framework~\cite{yuan2020plug}. 

Different from previous work~\cite{Yuan2021_TPAMI_PnP}, hereby we tend to use a lightweight learning method, e.i., the KSVD dictionary learning algorithm as the denoising prior.  
In the following, we briefly introduce the KSVD algorithm.

To be concrete, a noisy image $\mathbf{Q}$ can be represented as follows,
\begin{align}
    \mathbf{Q} = \mathbf{X} + \mathbf{Z},
\end{align}
where $\mathbf{X}$ denotes the original clean image that we want to obtain and $ \mathbf{Z}$ is the noise that we aim to remove; hereby, we abuse the notation of $\mathbf{X}$ to show both the clean images and videos in the SCI forward model without confusion.
The only information we have is the noisy observation  $\mathbf{Q}$. 

Under the assumption that by dividing the $\mathbf{X}$ into small (overlapping) patches, each (vectorized) patch 
$\mathbf{x}_i \in {\mathbb R}^{p^2}$ (with $p\times p$ the patch size and usually $p=8$) could be represented by an over-complete matrix $\mathbf{D}$ which we call the dictionary, the problem turns to an optimization problem  \cite{elad2006image} as follows,
 \begin{align}
 \label{eq:opt3}
 {\mathbf{x}}_i = {\mathbf{D}}{\mathbf{\alpha}}_i,  \quad  \mathbf{\hat\alpha}_i = \arg\min_{\mathbf{\alpha}_i}||\mathbf{{q}}_i-\mathbf{{D}}{\mathbf{\alpha}}_i||_2^2 + \mu||\alpha_i||_{0},  ~\forall i=1, \dots N_p,
 \end{align}
 where ${\alpha}_i$ represents the sparse coefficients of the patch $ {\mathbf{x}}_i$. $\mathbf{{q}}_i$ is the corresponding (vectorized) noisy observation,  ${\mu}$ is a weighting parameter and $N_p$ is the total number of patches. Solving norm-minimization equation with constraint condition like Equation \eqref{eq:opt3} gives the estimation ${\hat{\alpha}}_i$. Specifically, given the dictionary $\mathbf{D}$, ${\hat{\alpha}}_i$ can be solved by the orthogonal matching pursuit (OMP) algorithm \cite{Donoho12OMP}.
 
Regarding the dictionary $\mathbf{D}$, it needs to be trained by jointly considering all the image patches (in the training set) as these patches share the same dictionary.
Furthermore, $\mathbf{D}$ can be pre-trained or learned during the noise removal process.
Please refer to the KSVD paper \cite{elad2006image} for details.

By plugging this KSVD denoising prior into the denoising step \eqref{Eq:denoisePnP} in the PnP framework, we achieve the proposed PnP-KSVD algorithm for SCI reconstruction. 
In our work, we use the pre-trained $\mathbf{D}$ and conduct denoising in each frame of the SCI measurement.
An online dictionary learning or adaptively updating the dictionary during the reconstruction can lead to better results but takes a longer time. We leave this for future work.

%% file: experiment.tex
\section{experiment}
\subsection{Dataset}
In our experiments, we use simulation data including \textit{Traffic}, \textit{Runner}, \textit{Crash}, \textit{Kobe}~\cite{yang2014video}, and \textit{Drop} datasets. The original images are resized to $256 \times 256$ through down sampling. Following the setting in~\cite{liu2018rank}, eight $(B = 8)$ consequent frames are modulated by shifting binary masks and then collapsed to a single measurement.

\subsection{Compared method}
For both quantitative and qualitative evaluations, we compare our method with GAP-TV~\cite{Yuan16ICIP_GAP}, which solves the video SCI problem with a total variation minimization, and serves as a proper baseline for SCI reconstruction. 
Reconstruction results on the above-mentioned simulation datasets are taken into comparison. Since the number of measurements in each simulation data sets varies, we use the mean value across all testing trials to evaluate the reconstruction performance. For a comprehensive comparison, we use two image quality metrics, peak signal to noise ratio (PSNR) and structural similarity index (SSIM)~\cite{Wang04imagequality}.


\begin{figure}[!h]
    \centering
  \includegraphics[width=\textwidth]{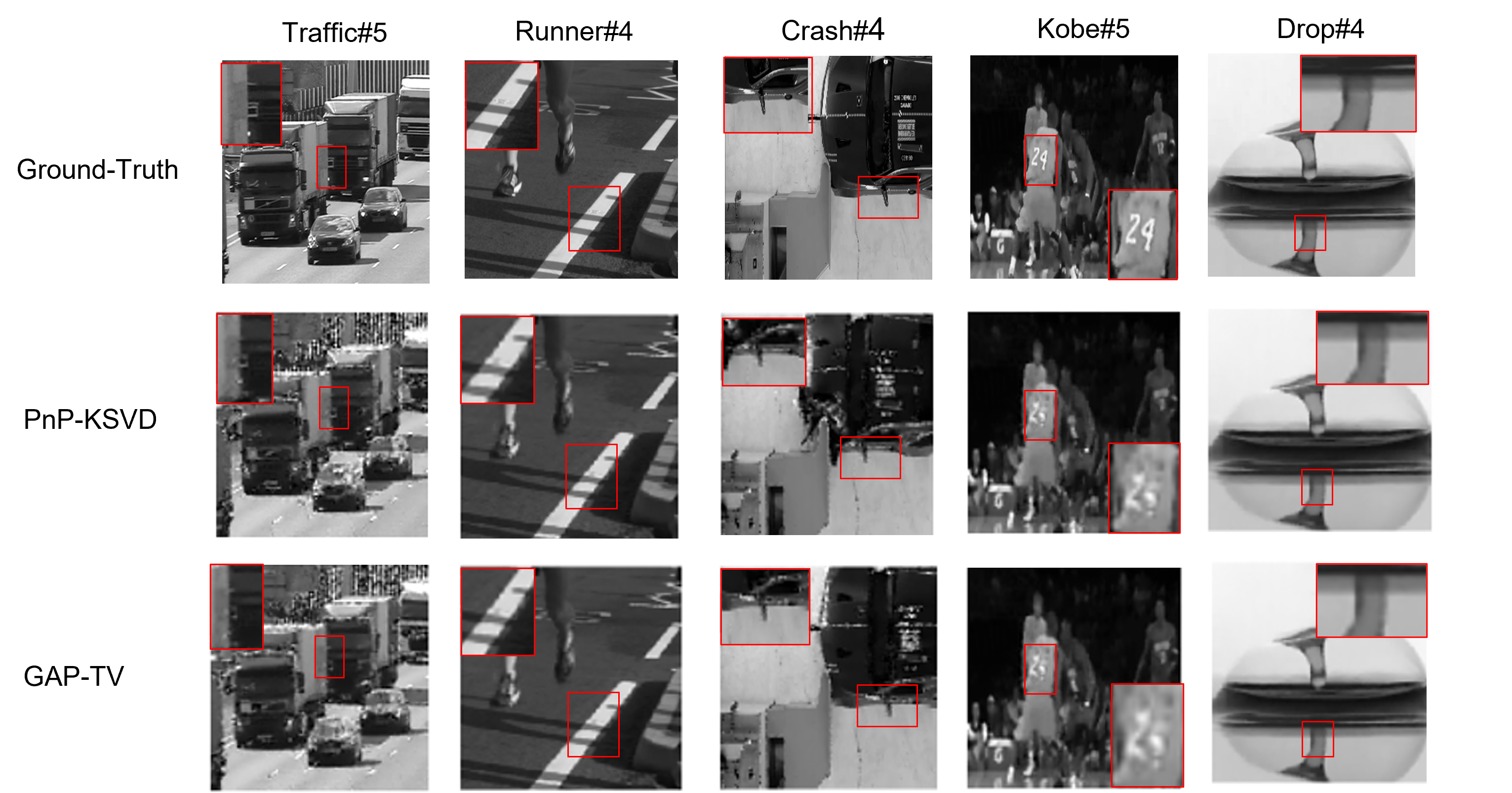}
  \caption {SCI reconstruction results comparison. Using PnP-KSVD algorithm and GAP-TV algorithm to reconstruct videos named \textit{Traffic}, \textit{Runner}, \textit{Crash}, \textit{Kobe}, \textit{Drop} respectively. We could see that PnP-KSVD performs better than GAP-TV in reconstruction qualitatively. }
  \label{fig:comparison}
\end{figure}

\begin{figure}[htbp]
\centering
\subfigure[]{
\begin{minipage}[t]{\textwidth}
\centering
\includegraphics[width=0.8\textwidth]{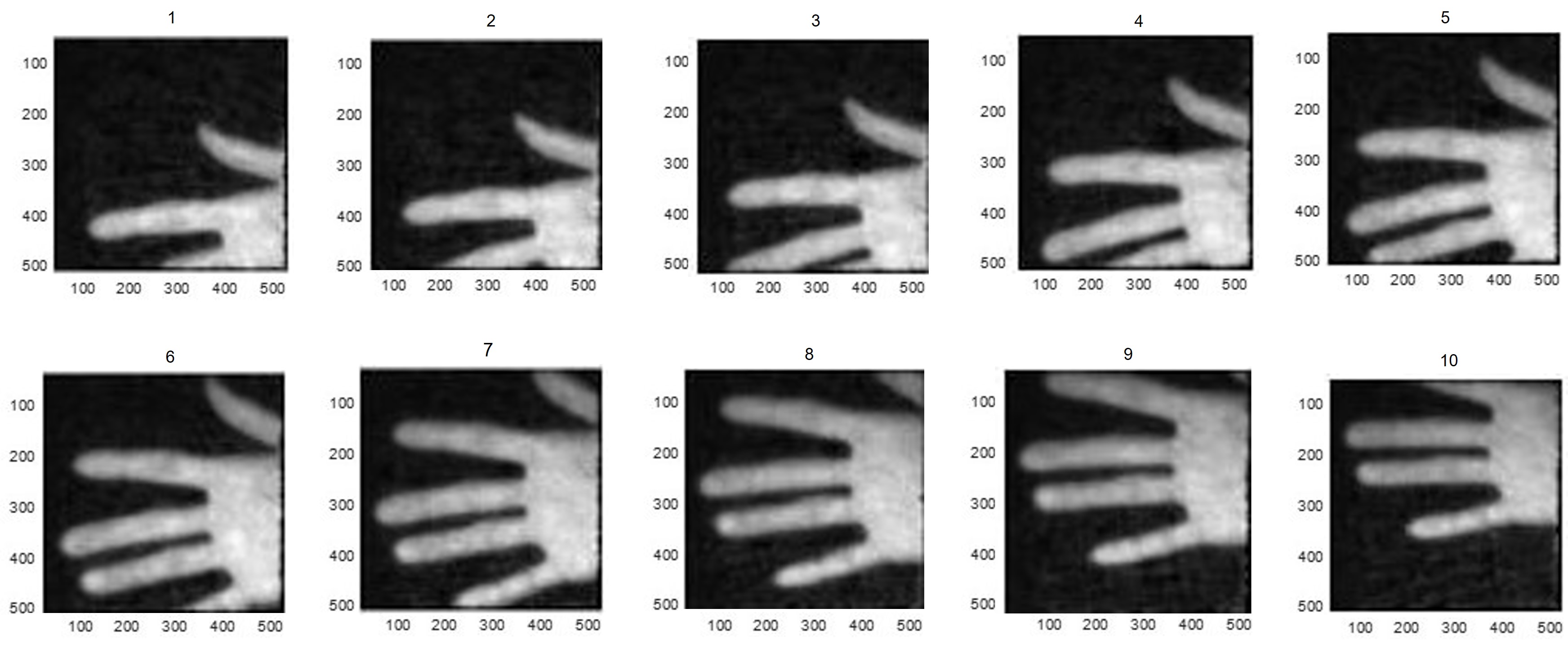}
\label{fig:comparison1}
\end{minipage}%
}%

\subfigure[] 
{\begin{minipage}[t]{\textwidth}
\centering
\includegraphics[width=\textwidth]{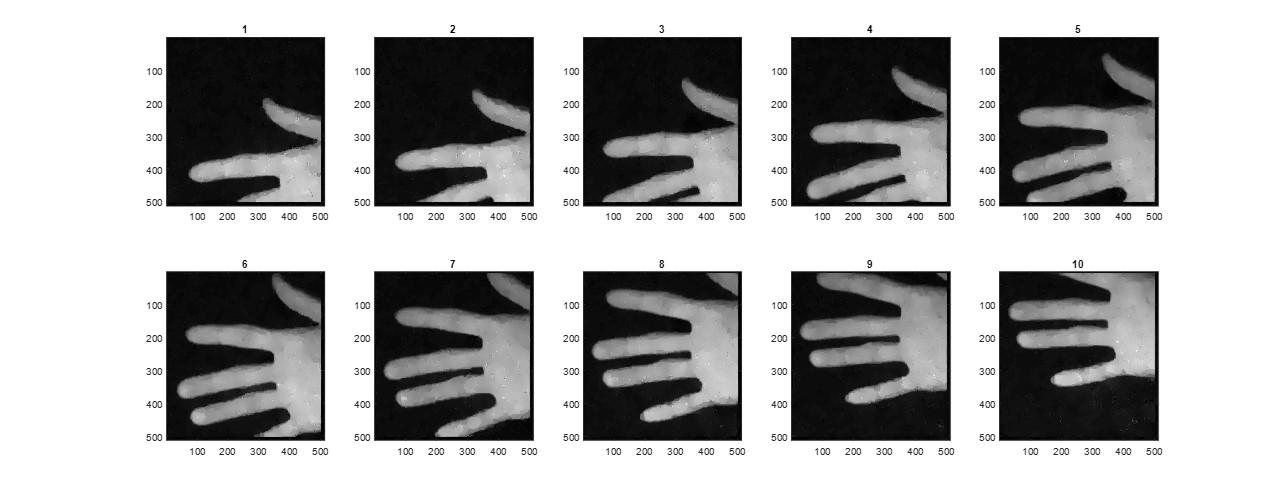}
\label{fig:comparison2}
\end{minipage}%
}%
\centering
\caption{Real data results. (a) and (b) show real-world reconstruction using GAP-TV and PnP-KSVD, respectively. Results in (a) are more blurry and noisy in comparison to those in (b).}
\end{figure}



\begin{table}[ht]
\caption{Performance comparison of PnP-KSVD and GAP-TV on four simulation scenes using PSNR (in dB, left in each cell) and SSIM (right in each cell) as the metrics. Mean values of PSNR and SSIM corresponding to experimental results are shown as follows. We could see that PnP-KSVD performs better than GAP-TV under metrics of PSNR and SSIM quantitatively.} 
\label{tab:Paper Algorithm}
\begin{center}       
\begin{tabular}{|l|l|l|l|l|l|} 
\hline
\rule[-1ex]{0pt}{3.5ex}  Algorithm & Traffic & Runner  & Kobe & Drop & Average  \\
\hline
\rule[-1ex]{0pt}{3.5ex}  PnP-KSVD & 21.715, 0.801 & 33.257, 0.925 & 30.976, 0.888 & 39.703, 0.974 & 31.413, 0.897 \\
\hline 
\rule[-1ex]{0pt}{3.5ex} GAP-TV  & 20.800, 0.707 & 30.651, 0.921 & 28.847, 0.895 & 36.157, 0.976 & 29.114, 0.875   \\ 
\hline 
\end{tabular}
\end{center}
\end{table}

\subsection{Experiment results}
Examining Table \ref{tab:Paper Algorithm}, our method outperforms GAP-TV considerably on various datasets, including \textit{Traffic}, \textit{Runner}, \textit{Kobe}, \textit{Drop}. And our mean values of PSNR and SSIM across all the testing trials are also superior to that of GAP-TV.
Figure~\ref{fig:comparison1} and Figure~\ref{fig:comparison2} presents the visual comparisons of real data results. In comparison to GAP-TV, our method can provide finer details.

%% file: conclusion.tex
\section{conclusion}

In this paper, we revisit the kernel singular value decomposition (KSVD), a dictionary learning method, and apply it for SCI reconstruction. By integrating KSVD-based denoiser into the PnP framework, we provide a new baseline for video SCI reconstruction. Experiments on various datasets show that with a well-trained dictionary, the results of our proposed PnP-KSVD are better than GAP-TV. In addition, our method retains a good trade-off between quality and training difficulty.